# Harassment detection: a benchmark on the #HackHarassment dataset

Alexei Bastidas, Edward Dixon, Chris Loo, John Ryan
Intel
e-mail: edward.dixon@intel.com

## Introduction

*Online harassment has been a problem to a greater or lesser extent since the early days of the internet. Previous work has applied anti-spam techniques like machine-learning based text classification (Reynolds, 2011) to detecting harassing messages. However, existing public datasets are limited in size, with labels of varying quality. The #HackHarassment[1] initiative (an alliance of tech companies and NGOs devoted to fighting bullying on the internet) has begun to address this issue by creating a new dataset superior to its predecssors in terms of both size and quality. As we (#HackHarassment) complete further rounds of labelling, later iterations of this dataset will increase the available samples by at least an order of magnitude, enabling corresponding improvements in the quality of machine learning models for harassment detection. In this paper, we introduce the first models built on the #HackHarassment dataset v1.0 (a new open dataset, which we are delighted to share with any interested researcherss) as a benchmark for future research.*

## Related Work

Previous work in the area by Bayzik 2011 showed that machine learing and natural language processing could be successfully applied to detect bullying messages on an online forum. However, the same work also made clear that the limiting factor on such models was the availability of a suitable quantity of labeled examples. For example, the Bayzick work relied of a dataset of 2,696 samples, only 196 of which were found to be examples of bullying behaviour. Additionally, this work relied on model types like J48 and JRIP (types of decision tree), and k-nearest neighbours classifiers like IBk, as opposed to popular modern ensemble methods or deep neural-network-based approaches.

## Methodology

Our work was carried out using the #HackHarassment Verison 1 dataset, the first iteration of which consists exclusively of Reddit posts. An initially random selection of posts, in which harassing content occured at a rate of between 5% and 7% was culled of benign content using models training on a combination of existing cyberbullying datasets (Reynolds 2001, also "Improved cyberbullying detection through personal profiles). Each post is labelled independently by at least five Intel Security Web Analysts.  (a post is considered "bullying" if it labelled as such by 20% or more of the human labelers - as shown in the following histogram, a perfect consensus is relatively rare, and so we rate a post as "harassing" if 20% - 2 of our 5 raters - consider it to be harassing).  This is a relatively balanced dataset, with 1,280 non-bullying/harassing posts,, and 1,118 bullying/harassing examples.

---

[1] "Hack Harassment." 2016. 26 Jul. 2016 <http://www.hackharassment.com/>

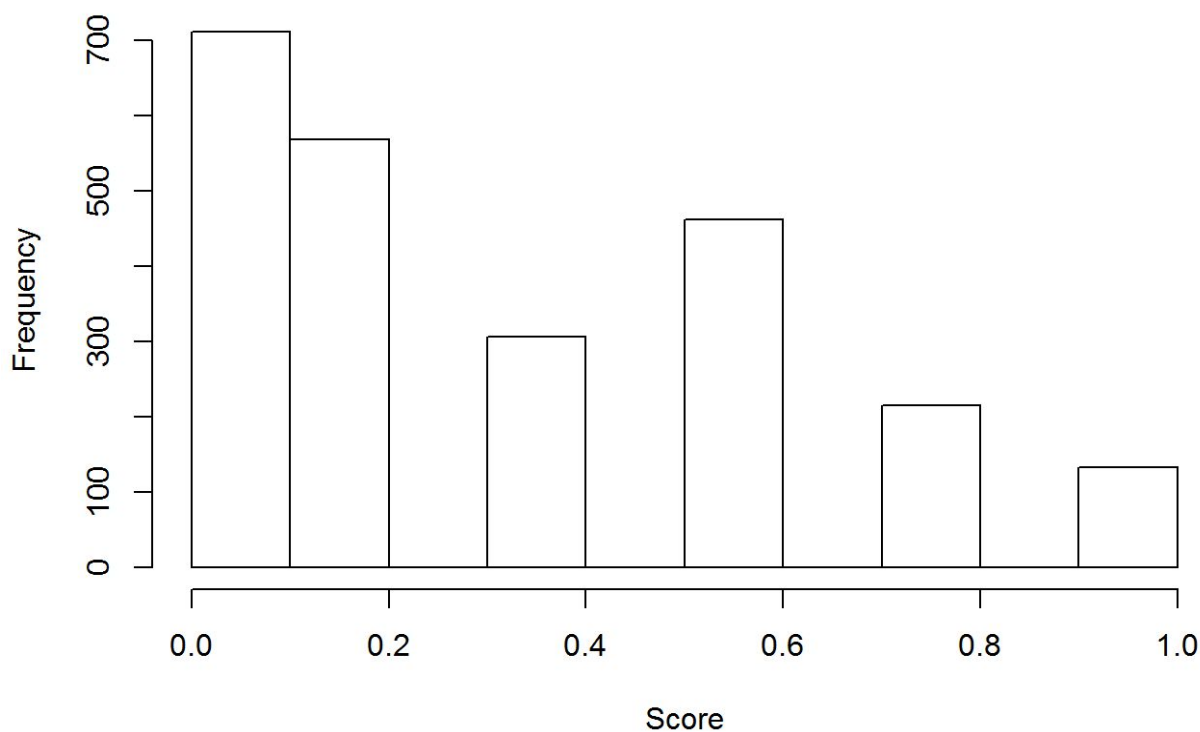

All pre-processing, training and evaluation was carried out in Python, using the popular SciKit-Learn library[2] (for feature engineering and linear models) in combination with Numpy[3] (for matrix operations), Keras[4] and TensorFlow[5] (for models based on deep neural networks - DNNs).

For the linear models, features were generated by tokenizing the text (breaking it aparting into words), hashing the resulting unigrams, bigrams and trigrams (collectiojns of one, two, or three adjacent words) and computing at TF/IDF for each hashed value. The resulting feature vectors were used to train and test Logistic Regressioin, Support Vector Machine and Gradient Boosted Tree models, with 80% of data used for training and 20% held out for testing (results given are based on the held-out 20%).

For the DNN-based approach, a similar approach was taken to tokenization, both bigram and trigram hashes were computed; these were one-hot encoded, and dense representations of these features were learned during training, as per Joulin 2016.

---

[2] "scikit-learn: machine learning in Python — scikit-learn 0.17.1 ..." 2011. 29 Jul. 2016 <http://scikit-learn.org/>
[3] "NumPy — Numpy." 2002. 29 Jul. 2016 <http://www.numpy.org/>
[4] "Keras Documentation." 2015. 29 Jul. 2016 <http://keras.io/>
[5] "TensorFlow — an Open Source Software Library for Machine ..." 2015. 29 Jul. 2016 <https://www.tensorflow.org/>

The FastText model used is a python implenmentation of the model described in "Bag of Tricks for Efficient Text Classification."[6] . For the text encoding, bigrams and trigrams are used. 20% of the data was held out for testing.

The Recurrent Character Level Neural Network model consists of 2 GRU layers of width 100 followed by a Dense Layer of size 2 with softmax on the output, Between each of the layers batch normailzation is performed. The optimiser used was rmsprop. For data preperation each of characters was onehot encoded and each sample was truncated/padded to 500 charcters in length. 20% of the data was held out for testing.

## Results

| Model | Precision (Harassing) | Recall (Harassing) |
|---|---|---|
| Gradient Boosted Trees (Scikit-Learn) | 0.80 | 0.71 |
| Bernoulli Naive Bayes | 0.54 | 0.30 |
| FastText | 0.60 | 0.78 |
| Recurrent Character Level Neural Network | 0.71 | 0.73 |

## Conclusions

We have presented the first results on a new open cyberbullying/harassment dataset. While our models clearly demonstrate a degree of ability to discriminate between the content classes, the achieved precision in particular falls far short of our ambitions for #HackHarassment.

Over the coming months, we'll massively expand the size of our labelled dataset, and review our labelling methodology, anticipating that a larger dataset will facilitate more accurate models. We look forward both to the availability of a larger dataset, and to seeing the development of classifiers that improvement on our work, and welcome partners able to contribute either in terms of expanding the dataset or improving the modelling.

---

[6] "fastText" 2015. 22 Jul. 2016 <https://github.com/sjhddh/fastText>